\documentclass[preprint,12pt]{elsarticle}

\usepackage{graphicx}
\usepackage{subfigure}

\usepackage[latin1]{inputenc}
\usepackage{amssymb}
\usepackage{amsthm}
\usepackage{amsmath}
\usepackage{colortbl}
\usepackage{enumerate}
\usepackage{exscale}
\usepackage{relsize}
\usepackage{algorithm}
\usepackage{algorithmicx}
\usepackage{algpseudocode}
\usepackage{array}
\usepackage{natbib}
\usepackage{booktabs}
\usepackage{upgreek}

\definecolor{light}{gray}{.85}

\newcommand{\Rmnum}[1]{\expandafter\@slowromancap\romannumeral #1@}

\biboptions{compress}
\biboptions{}
\journal{arxiv.org}

\begin{document}
\begin{frontmatter}


% use the tnoteref command within \title for footnotes;
% use the tnotetext command for the associated footnote;
% use the fnref command within \author or \address for footnotes;
% use the fntext command for the associated footnote;
% use the corref command within \author for corresponding author footnotes;
% use the cortext command for the associated footnote;
% \author{Name\corref{cor1}\fnref{label2}}
% \address{Address\fnref{label3}}

\title{MSA-MIL: A deep residual multiple instance learning model based on multi-scale annotation for classification and visualization of glomerular spikes}
% \tnotetext[t1]{ }
% \author[label1,label2]{<author name>}
% \address[label1]{<address>}
% \address[label2]{<address>}

\author[author1]{Yilin Chen}
%\ead{chenyilin100@gmail.com}
\author[author1]{Ming Li\corref{cor1}}
\ead{liming01@tyut.edu.cn}
\author[author1,author2]{Yongfei Wu}
%\ead{yongfeiwu522@sina.com}
\author[author1]{Xueyu Liu}
%\ead{liuxueyu1122@gmail.com}
\author[author1]{Fang Hao}
%\ead{haofang12358@163.com}
\author[author1]{Daoxiang Zhou}
%\ead{zhoudaoxiang@tyut.edu.cn}
\author[author3]{Xiaoshuang Zhou}
%\ead{xiaoshuangzhou66@163.com}
\author[author4]{Chen Wang\corref{cor1}}
\ead{wangchen8877322@163.com}

\address[author1]{College of Data Science, Taiyuan University of Technology, Taiyuan, Shanxi, China}
\address[author2]{Department of Computer and Information Science, Faculty of Science and Technology, University of Macau, Taipa, Macau, China}
\address[author3]{Department of Nephrology, Shanxi Provincial People's Hospital, Taiyuan, Shanxi, China}
\address[author4]{Department of Pathology, Second Hospital of Shanxi Medical University, Taiyuan, Shanxi, China}

\cortext[cor1]{Corresponding author.}

\begin{abstract}
Membranous nephropathy (MN) is a frequent type of adult nephrotic syndrome, which has a high clinical incidence and can cause various complications. In the biopsy microscope slide of membranous nephropathy, spikelike projections on the glomerular basement membrane is a prominent feature of the MN. However, due to the whole biopsy slide contains large number of glomeruli, and each glomerulus includes many spike lesions, the pathological feature of the spikes is not obvious. It thus is time-consuming for doctors to diagnose glomerulus one by one and is difficult for pathologists with less experience to diagnose. Automatical classification technologies assisting the clinical diagnosis of glomerular spikes are urgently needed to leverage the deficiency of traditional diagnostic methods. In this paper, we establish a visualized classification model based on the multi-scale annotation multiple instance learning (MSA-MIL) to achieve glomerular classification and spikes visualization. The MSA-MIL model mainly involves three parts. Firstly, U-Net is used to extract the region of the glomeruli to ensure that the features learned by the succeeding algorithm are focused inside the glomeruli itself. Secondly, we use MIL to train an instance-level classifier combined with MSA method to enhance the learning ability of the network by adding a location-level labeled reinforced dataset, thereby obtaining an example-level feature representation with rich semantics. Lastly, the predicted scores of each tile in the image are summarized to obtain glomerular classification and visualization of the classification results of the spikes via the usage of sliding window method. The experimental results confirm that the proposed MSA-MIL model can effectively and accurately classify normal glomeruli and spiked glomerulus and visualize the position of spikes in the glomerulus. Therefore, the proposed model can provide a good foundation for assisting the clinical doctors to diagnose the glomerular membranous nephropathy.\\
\end{abstract}


\end{frontmatter}

% \linenumbers
\newpage
\section{Introduction}
\label{sec-intro}
\begin{figure}[htb]
\centerline{\includegraphics[width=6 in,height=5.76 in]{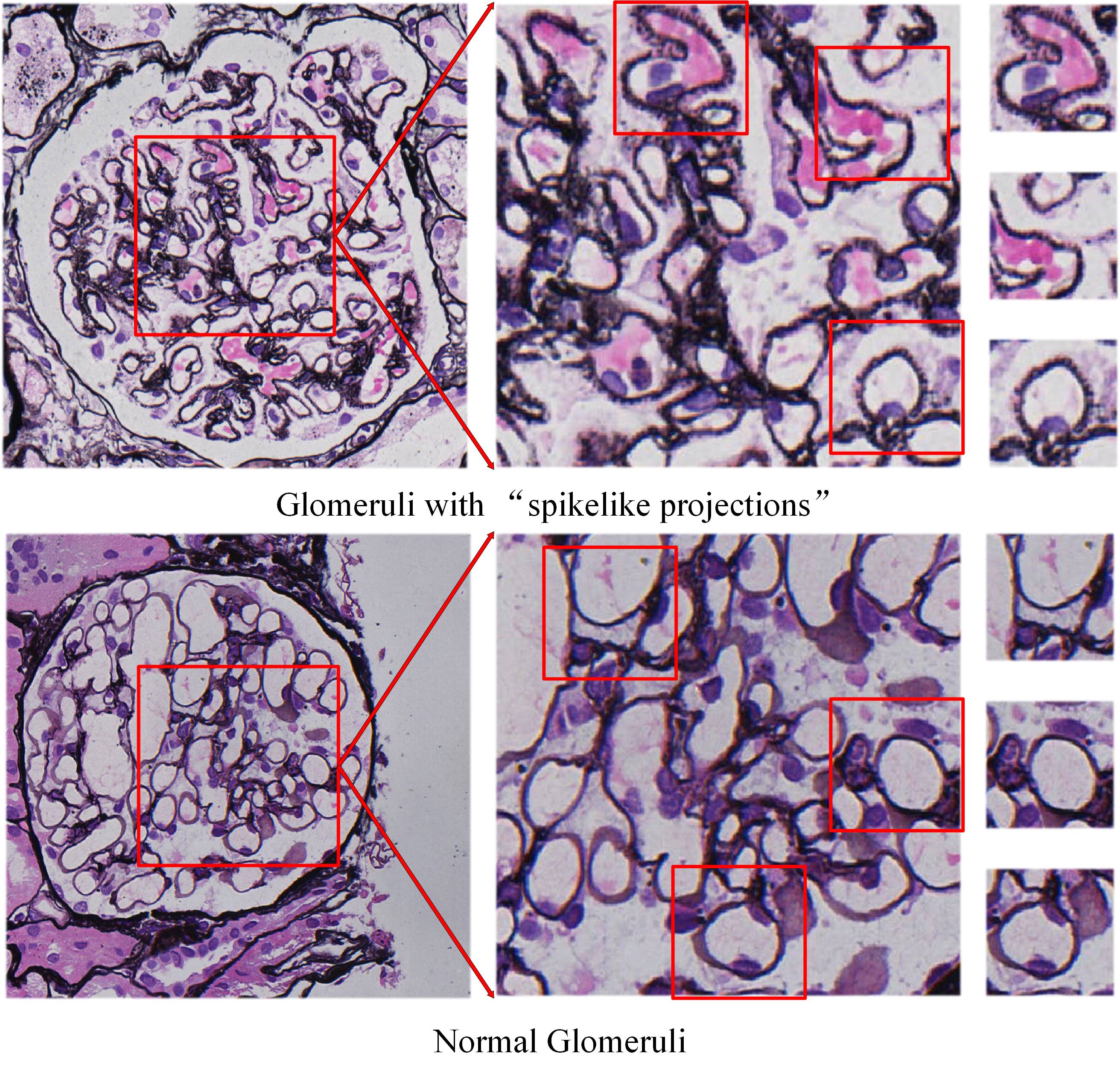}}
\caption{Illustration of glomeruli with or without spikelike projections. Top row: Glomeruli with spikelike projections; Bottom row: Normal glomeruli without spikelike projections.}
\label{Schematic_diagram_of_spikes}
\end{figure}

In recent years, the aging problem of population is becoming more and more serious in the world, and the incidence rate of nephropathy has been always increasing. According to relevant surveys, the incidence rate of chronic kidney disease in China is about 11$\%$, and about 130 million patients were diagnosed. Membranous nephropathy (MN) is a common type of adult nephrotic syndrome, which frequently occurs in patients over 40 years old. Related research indicates that the clinical incidence rate of it accounts for more than 10$\%$ of glomerular diseases. Pathological biopsy as the gold standard plays a pivotal role for the diagnosis of membranous nephropathy. The pathological change of membranous nephropathy is characterized as that a large number of immune complex sediments can be observed on the epithelial side of glomerular capillary loop, and reactive proliferation of basement membrane can be seen between sediments, which is called spikelike projections (Referred to as spikes in this paper). In Fig.\ref{Schematic_diagram_of_spikes}, we give an illustration of the glomerular basement membrane with or without spikes. As shown in Fig.\ref{Schematic_diagram_of_spikes}, The outside of the glomerular basement membrane with spikes is not smooth, while the outside of the glomerular basement membrane without spikes is smooth.

The spikes on the glomerular basement membrane is a prominent pathological feature in the diagnosis of MN. However, due to the whole biopsy slide contains large number of glomeruli and each glomerules also includes many spike lesions, resulting the pathological feature of the spikes is not obvious. The traditional method of pathological diagnosis requires the pathologist to look for the spikes one by one under the microscope and make a diagnosis for whole glomeruli, it is time-consuming for pathologists to diagnose the whole biopsy slide and is more difficult for pathologists with less experience to diagnose. At the meantime, the subjective difference of pathological diagnosis, fatigue reading and other factors are inevitably lead to a certain rate of misdiagnosis. Besides, Other factors involving the long training cycle of pathologists, the high risk of work and the low labor reward bring about the serious shortage of pathologists. Therefore, it is urgently necessary to design automatic classification method to assist the clinical doctors in the clinical diagnosis of renal pathology.

Digital pathology is a research field that transforms pathological sections into digital pathological images and carries out diagnosis and quantification. Its rapid development provides the possibility to address the problem of diagnosing the MN. The utilization of computer aided diagnosis technology greatly improves the efficiency and accuracy of pathologists. At present, deep learning methods have been proposed in the literature and widely used in the auxiliary diagnosis of diseases. In the field of pathology, the diagnosis and segmentation of cancer regions have been studied a lot and prolific models and results are proposed \cite{Spanhol22016, Litjens2016, Janowczyk2016}. However, the research of digital pathology in the field of renal pathological image is not common, and its main research results are focused on the recognition of cell images with simple tissue structure (such as relatively simple background, relatively loose cell distribution, etc.), the detection and extraction of glomeruli \cite{Pedraza2017, Simon2018, Kawazoe2018} and the segmentation of glomeruli \cite{LiuR2019, Bueno2020}. To the best of our knowledge, only some research works have been proposed to tackle the classification task of specific glomeruli, Barros et al. \cite{Barros2017} constructed a model to classify proliferative lesions. Marsh et al. \cite{Marsh2018} and Kannan et al. \cite{Kannan2019} reported models to distinguish between sclerotic  and  nonsclerotic  glomeruli. Ginley et al. \cite{Ginley2019} constructed a model to classify diabetic nephropathy. Uchino et al. \cite{Uchino2020} constructed a model to classify seven pathological findings in glomerular images using deep learning by fine-tuning of convolutional neural network. Chagas et al. \cite{Chagas2020} constructed a model used convolutional features and support vector machine to classify glomerular hypercellularity. Since the spikes are fine-grained and the difference between the lesion and the background is not significant, there is few study on the classification of spikes glomeruli so far.

The diagnosis of glomerular spikes can be intelligently implemented by the classification of normal glomeruli and spiked glomeruli. Most existing classification methods employ deep learning based methods to build classification models via the data-driven manner. This type of classification method directly extracts the features of the entire image, and cannot learn the fine-grained lesion features. Therefore, the classification results are difficult to meet the requirements of clinical assistant diagnosis in the pathological image field. In addition, the current classification methods based on the deep learning method are lack of interpretation, and the classification results of the model are invisible. In other words, the doctor cannot visually confirm whether the classification result is correct. This shortcoming therefore causes the doctor to distrust the algorithm, making the designed model cannot be successfully applied to clinical diagnosis.

As a remedy to overcome the above mentioned defects, this paper proposes a visualized classification method based on multi-scale annotations multiple instance learning (MSA-MIL), which could achieve the classification of normal glomeruli and spiked glomeruli and the visualization of the classification results. The proposed model exploit multiple instance learning as a training framework combined with U-Net and ResNet. the MSA-MIL model mainly includes three phases. We firstly use the U-Net to pre-extract the region of the glomeruli such that the learned features by the succeed algorithm are focused on the glomeruli itself. Then, we feed the MIL trained data into the ResNet network to obtain an instance-level classifier. In the network training process, the entire image is regarded as a bag and the label of the entire image is denoted as the label of the bag, the extracted region of glomeruli is divided into many tiles with each tile in the region is regarded as an instance and the label of it is denoted as the label of the instance. Besides, we use a multi-scale annotation method to add location-level labeling reinforced data to enhance the learning ability of the network, thereby obtaining an example-level feature representation with abundant semantic information and improving the classification accuracy caused by unstable labels in multiple instance learning. Finally, the predicted scores of each tile in the image are summarized up to obtain glomerular classification result and give the visualized heatmap of spikes via the usage of sliding window method. The experimental results showed that the average F-measure value of the MSA-MIL is $0.9580$ and the average time of diagnosing a glomerulus image is $1.33s$, which obtains better performance compared with other classification networks and accelerates the screening of slide image, enhancing the clinical diagnosis. Furthermore, the MSA-MIL can give the visualized location of the spikes which could supplies trustable classification results for clinical doctors. Therefore, the proposed MSA-MIL can provide a good foundation for assisting the clinical doctors to diagnose the glomerular membranous nephropathy.

\section{Related Works}
\label{sec-relatedworks}
\subsection{Deep neural network}
\label{subsec-deep network}
Recently, deep learning technology as a class of multi-layer neural network learning algorithms can automatically learn hierarchical feature representations of the input data and form higher abstract features (attribute categories or features) by combining low-level features through deep nonlinear network structures. At present, deep learning methods have been widely used in the auxiliary diagnosis of diseases, and have made great progress in handling CT images \cite{Shen2015, Setio2016, YuQ2019}, MR images \cite{Dou2016, Yang2015, Aldoj2019}, retinal fundus images \cite{Gulshan2016} and skin images \cite{Esteva2017}. Among the available deep learning methods, deep convolution neural network (DCNN) and convolution residual network (CRN) are the most popular algorithm.

In the field of pathological images classification, especially for the breast cancer pathology classification, deep learning has shown good performance in dealing with the breast cancer data. Spanhol et al. \cite{Spanhol2016} used the AlexNet network to classify breast cancer, and the recognition rate obtained was 6$\%$ higher than that of traditional algorithms, but the accuracy remains low and could not conform to the clinical requirements. Wei et al. \cite{Wei2017} used GoogleNet as the basic network architecture and used the labels of classes and subclasses as a priori knowledge to construct the breast image CNN (BICNN) model. Although this model arrives very high accuracy of classification, network deepening lead to the degraded efficiency. The ResNet \cite{He2016} proposed in $2015$ is currently the most widely used CNN feature extraction network, which can be used for classification. It avoids the problem of network degradation with the number of network layers increasing. The ResNet introduces the residual network structure, which can deepen the network depth and achieves better classification effect. However, due to the specificity of renal pathological lesions, the classification methods that directly extract the features of the entire image cannot learn the fine-grained lesion features. Therefore, the classification results are not acceptable and can not apply to clinical assistant diagnosis in the pathological image field. Additionally, the current classification methods are lack of interpretability, and the classification results of the model are invisible. Thus these classification models cannot convince the clinical doctors. For the segmentation of pathological images, the U-Net proposed by Ronneberger et al. \cite{Ronneberger2015} is the most widespreadly used segmentation framework based on CNN backbone and is one of the most influential models for medical image segmentation because of its small parameters and good segmentation effect. The U-Net combines the same number of up-sampling and down-sampling layers, which can generate a segmented image from the entire image by the forward algorithm in a fashion of end to end. At the meantime, the skip connections across the upper and lower sampling layers ensure the gradual preservation of information. When this network was first proposed, it was utilized for cell wall segmentation. Later, it has excellent performance in the detection of lung nodules \cite{Cao2019} and the extraction of blood vessels on the fundus retina \cite{Xian2018}.

\subsection{Multiple instance learning}
\label{subsec-MIL}
Multiple instance learning \cite{Dietterich1997} (MIL) is a weakly supervised learning method, which builds a classifier for multiple instance bags with classification labels and predicts unknown multiple instance bags and can realize the visualization of classification results without increasing the labeling cost. In MIL, the training set consists of a set of bags with classification labels, and there are several instances without classification labels in each bag. If the bag contains at least one positive instance, the bag is marked as a positive bag. If all instances of a bag are negative instances, the bag is marked as a negative bag. There are some researches in the literature which regard the problem of image classification as the problem of multiple instance learning \cite{Maron1997,Maron1998,Yang2000}. In recent years, MIL has been successfully used in pathological image classification \cite{Xu2016}. Yan et al. employed the MIL framework to complete the classification of colon cancer image by regarding the pathological image as a bag and dividing the image into multiple patches as an instance \cite{Xu2014}. Hou et al. \cite{Hou2015} Proposed a multiple instance convolution algorithm based on EM algorithm to classify gliomas. Campanella et al. \cite{Campanella2019} proposed a deep learning system based on multiple instance learning, which only uses the diagnosis report as the label of training, and classifies the pathological images of prostate cancer, basal cell cancer and breast cancer transferred to axillary lymph nodes. However, the label of the positive bag is not the real label of instance, the stability of the label is greatly challenged, which leads to incorrect classification results to a certain extent.

\section{Methods and model}
\label{methods}
\subsection{Multi-scale annotation}
\label{subsec-msa}
\begin{figure}[htb]
\centerline{\includegraphics[width=6 in,height=4.8 in]{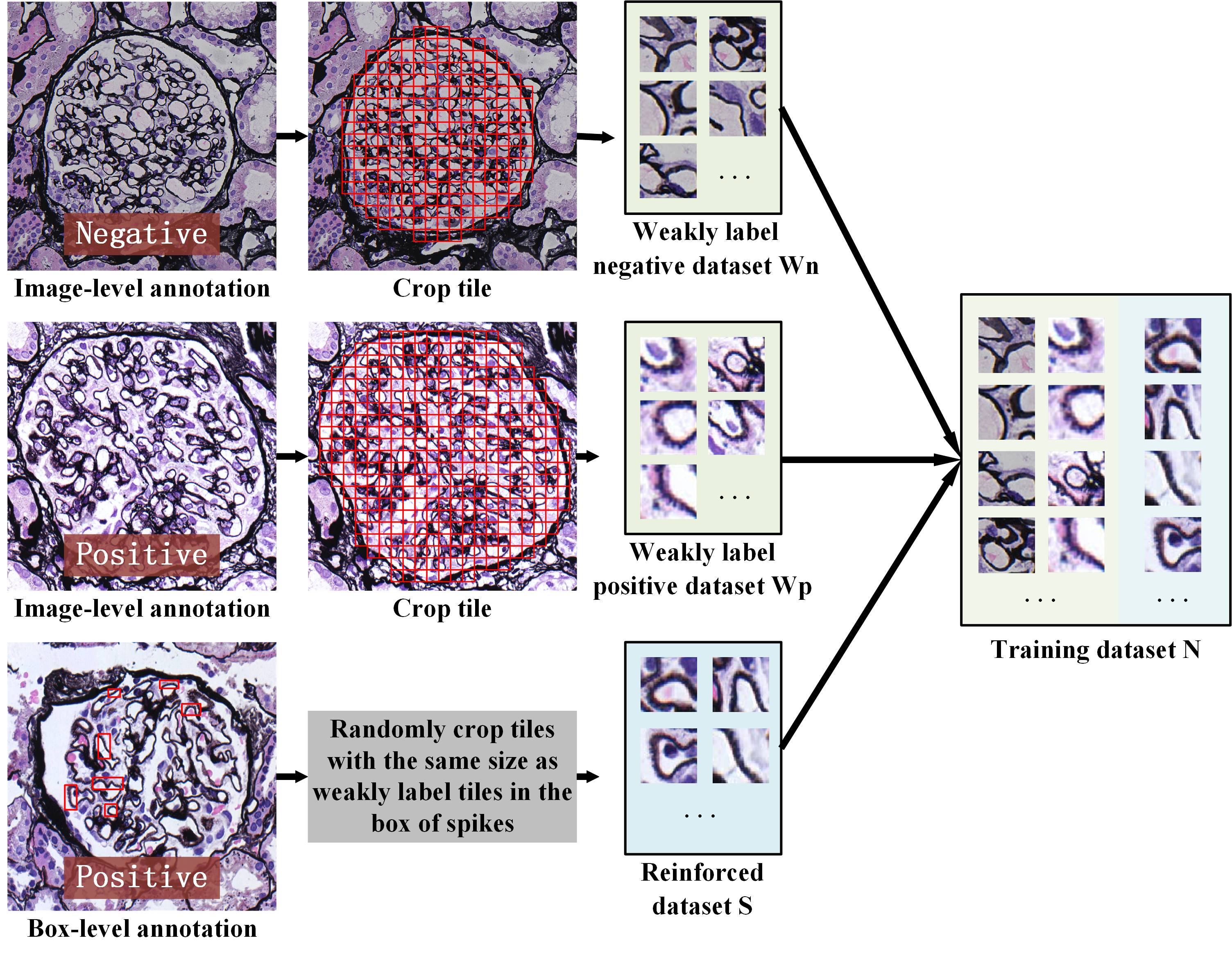}}
\caption{Demonstration of multi-scale annotation method. Top row: image-level annotation of negative image; Middle row: image-level annotation of positive image; Bottom row: location-level annotation of positive image.}
\label{Schematic_diagram_of_MSA}
\end{figure}

In this work, we propose a multi-scale annotation (MSA) method which is a data processing method used for enhancing the learn capacity of MIL. It overcomes the defect of unstable labels by adding instance-level labels. Specially, the MSA hires image-level annotation and location-level annotation (box-level annotation) to annotate images, respectively. For the image-level annotation, the pathologist gives a label of the whole images when cropping the image into tiles and each label of the tile is deemed as the label of whole image. Suppose $x$ denotes a tile, $y$ represents the label of the tile, $i$ is bag index, $j$ is tile index. For the negative image, we obtain weakly negative labeled data set $W_n(x_w^{(ij)},y_w^{(ij)})$. For positive image, we get weakly positive labeled data set $W_p(x_w^{(ij)},y_w^{(ij)})$. While for the location-level annotation, the pathologist annotates the true location of glomerular basement membrane spikes with a box in the glomerulus and gets the reinforced data set $S(x_s^{(ij)},y_s^{(ij)})$.  Due to the location-level annotation only labels the location of spikes with a box, it is a reinforced data set of positives. Therefore, the MSA method adds a small number of instance-level real labeled data into training data set, strengthening the supervision information in the training process of instance-level classification networks. In Fig.\ref{Schematic_diagram_of_MSA}, we give an illustration of the MSA method display the MSA method involving the image-level annotation and the location-level annotation.

The weakly labeled training set $W_n(x_w^{(ij)},y_w^{(ij)})$, $W_p(x_w^{(ij)},y_w^{(ij)})$ and the reinforced data set $S(x_s^{(ij)},y_s^{(ij)})$ are gathered and formed the total training data set $N(x_N^{(ij)},y_N^{(ij)})$, It would be used to train the instance-level classification networks. The specific training method will be described in detail at the next section.

\subsection{MSA-MIL}
\label{subsec-model}
\begin{figure}[htb]
\centerline{\includegraphics[width=6 in,height=4.08 in]{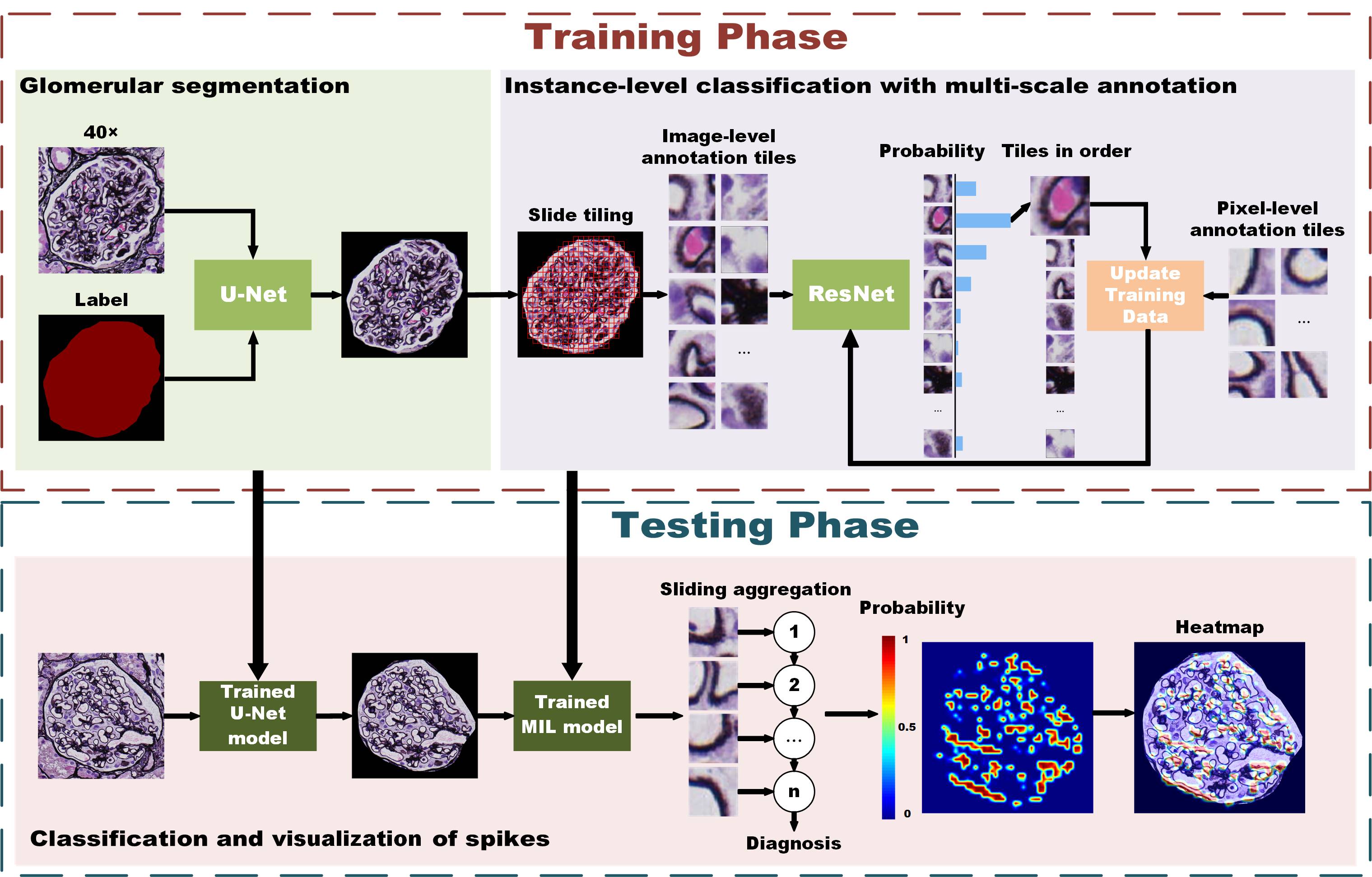}}
\caption{The architecture of classifying and visualizing Glomerular spikes proposed in this paper.}
\label{Structure of MSA-MIL}
\end{figure}

To solve the problems of instability of labels in MIL and non-visualization of classification results in the traditional classification methods, a MIL method based on multi-scale annotation is developed in this section to achieve the classification of glomerular spikes and visualization of the classification results. The MSA-MIL model mainly contains three phases. In the first phase, The U-Net is used to delineate the boundaries of the glomeruli, and some other digital image processing methods such as cavity filling and retention of connected domains are used to post-process the segmentation regions of the glomeruli. For the second phase, the obtained data by the U-Net is fed into the ResNet model under the MIL framework for training and a classifier is obtained. In the process of network training, we add a small amount of reinforced data via the MSA method for strengthening the learning capability of the network, so as to obtain instance level feature representation with rich semantic information. In the last phase, the predicted scores of each tile in the glomerular images are aggregated to get the classification results and visualization of spikes. The network architecture is shown in the Fig.\ref{Structure of MSA-MIL}.

\textbf{Segmentation of Glomerulus:} In the classification of spikes glomeruli, Since the major regions of concern is the glomerulus itself, the outside area of the glomerulus in the image can be precluded. The significant advantage of preprocessing the image is that the succeed learning algorithm mainly focuses on the important area making the algorithm, learns the characteristics of the glomerular internal lesions more effectively, and improves the accuracy of the classification algorithm. However, due to the internal structure of the glomeruli is complex, it is ineffective for traditional methods to segment the glomeruli. In this paper, we take U-Net as a data preprocessing step to segment glomerular edges because of its excellent segmentation performance at dealing with medical images. In order to ensure the stability of model training and optimize the effect of model training, we unify the size of glomerular image in the same range. Firstly, the glomerulus was randomly clipped from the WSIs magnified by $40\times$, and then the size of the glomerulus image was uniform $1100\times1100$, and the pixel was $0.25\upmu$m/pixel. The pathologist marks the outer edge of the glomeruli in the image. The original image and the masked image are put into the U-Net for training at the same time, in which the cross-entropy loss function is used.

\textbf{Instance-level classification with multi-scale annotation:} We resize the resolution of the segmented glomerular images as $1100 \times 1100$ for keeping the size of each image identical. We take images with spikes as positive bags and images without spikes as negative bags. For the images of positive and negative bag, we divide the extracted regions of them by U-Net into many tiles in which each tile is an instance in the bag, and the label of glomerular image is the label of tile in this image. According to the above analysis, the positive bag will contain at least one positive tile(with spikes), while the negative bag only contains the negative tile (without spikes). To ensure the features of spikes integral, we employ overlapping method in tile partition, and the overlapping step size is half of tile size. The selection of tile size is related to the size of specific lesions and would be discussed in experimental part.

By using the method of transfer learning, the classification training is first carried out on the open pathological data set, and the trained ResNet parameters and classifier $f$ are taken as the initial parameters and classifier of MSA-MIL training. In the training process, we use the image-level labeled tiles as the weakly labeled data set $W(x_w^{(ij)},y_w^{(ij)})$, and the location-level labeled tiles as the reinforced data set $S(x_s^{(ij)},y_s^{(ij)})$, where $x$ represents a tile, $y$ denotes the label of tile, $i$ is bag index, $j$ is tile index. The reinforced data $S$ set is stored in a temporary training set $D(x_d^{(ij,k)},y_d^{(ij,k)})$, where $k$ is the iterative index. For each bag, we first utilize initial classifier $f$ to class the data set $W$ and obtain the classification result of each tile in the bag, denoted as $Q_i=q_{(ij)}$. Then the $Q_i$ is processed by classifier $f$ and a probability vector $P_i=p_{(ij)}$ is obtained. Therefore, the tile with the highest positive probability can be expressed as $j=argmaxp_{(ij)}$. Take the positive instance with high probability in each positive bag, take the negative instance with high probability in each negative bag, and append them in the temporary training set $D(x_d^{(ij,k)},y_d^{(ij,k)})$. Finally, Using temporary training set $D$ to train the classification network and obtain a updated classifier $f$. By repeating the above three steps to iteratively fine-tune the network parameters, we get the final instance-level classification network for spikes.

In the process of updating classifier $f$, we use the cross-entropy loss by comparing the output $\hat{y}_i=f(q_{(ij)})$ with the target $y_i$ for each bag as follow:
\begin{equation} \label{Eq:loss1}
  {L_i=-y_ilog[\hat{y}_i]-(1-y_i)log[1-\hat{y}_i]}
\end{equation}
and the final loss is the weighted average of small batch loss.
\begin{equation} \label{Eq:loss2}
  {L_i= \sum_{n=1}^Ny_ilog[\hat{y}_i]+(1-y_i)log[1-\hat{y}_i]}
\end{equation}

Actually, the training of multiple instance learning in the network is to update the training set so as to finely tune the network. In multiple instance learning, there must be at least one positive instance classified by some classifiers. Instead, for a negative bag, all instances must be classified as negative instances. Given a bag, all the instances will be classified and sorted in detail according to its probability. If the bag is positive, the top sample positive probability should be close to 1; if the bag is negative, its positive probability should be close to 0.

\textbf{Classification and Visualization of spikes.} We use the sliding window aggregation method to classify spikes. By sliding a fixed size of window from top left to bottom right of glomerular image, we can obtain the classification probability for each window. The window can be regarded as a positive instance if the probability exceeds 0.5, and the window is classified as negative instance when the probability is less than $0.5$. For an input image, it is classified as positive category if it contains a window of positive instance, the image is considered as negative category when all windows are classified as negative instance.

As for images are classified as positive category (with spikes), we need further to give the specific location of tile with high probability of lesions such that the classified result is interpretable. After computing the classification probability of each window, we connect the probability of all of window and form the probability distribution matrix of the whole image of glomerulus. According to the probability distribution matrix of the spikes, we can obtain a thermogram of spikes. The final heatmap of glomerular spike can be drawn by superposing thermogram the with a certain transparency on the original image.

\section{Datasets and Results}
\label{sec-dataandresult}
The MSA-MIL model is based on MIL \cite{Campanella2019}. The proposed model is implemented on the PyTorch platform based on the Python environment running on a server with an NVIDIA Tesla K80 GPU. To achieve a more robust and accurate training for the proposed model, we train the model $30$ epochs with a batch size of $8$ and apply the Adam optimizer in which the learning rate base equal to $10^{-5}$ with downscaled factor of $0.3$. The $ResNet18$ model is pretrained based on BreaKHis dataset via the transfer learning and an initial instance-level classifier $f$ can be obtained, which has learned rich features of pathology and can directly applied to similar pathological images.

\subsection{Datasets}
\label{data}
The data of renal biopsy optical microscope slides are collected from the Second Hospital of Shanxi Medical University between $2014$ and $2019$ and the Shanxi Provincial People's Hospital between $2017$ and $2019$. The slides are stained with periodic acid-silver metheramine (PASM) and the slice is digitized as a whole slide image (WSI) by KF-PRO-005-EX digital slice scanner (resolution $0.25\upmu$m/pixel) with $40 \times$ objective and $10 \times$ eyepiece. All of the images are obtained by clipping the image with magnification of $400 \times$.

\textbf{Training dataset for segmentation.} Two pathologists first annotate and record the positions and coordinates of the glomerular in WSI by utilizing an image annotation system. Then, the pathologist cut the glomerulus as patch with the size of $1100 \times 1100$ pixels. Finally, The pathologist marked the edges of the glomerulus patch at the accuracy of pixel-level, and the obtained mask is served as the label of glomerulus segmentation. We apply Dice coefficient (DC) to determine that the labeled data is effective. If the DC value between two labeled data obtained from two pathologists is greater than $0.8$, then the two labeled data is effective and we choose one of them as the training data. Repeating the above operating procedure, $1015$ images of glomerular were labeled at pixel-level in which $914$ of them is choose as the training set and the remained $101$ images is used as the test set.

\textbf{Training dataset for classification.} Two pathologists annotate the glomeruli patch at the image level. The annotated results have three classes of positive (denoted as $+$), negative (denoted as $-$), unable to determine (denoted as $+,-$). In order to ensure the reliability of the data, only the glomerular patch diagnosed as the same category (negative or positive) by two pathologists is added in the training data set.
Additionally, to obtain reinforced training data, the glomerular image of positive category is annotated as a typical spikes by a box with $50 \times 50$ pixels at location-level. Finally, there are $1267$ images with $653$ positive glomeruli and $614$ negative glomeruli in the data set, and $250$ location-level labeled images are produced from $653$ positive glomeruli. We divide the dataset and select $252$ images as the test data set to evaluate the performance of the proposed model, and the remaining $1015$ images is as the training data set.

\subsection{Assessment criteria of proposed model}
To quantitatively evaluate the performance of the proposed model for classifying spiked glomeruli, we employ four metrics such as the Precision, the Recall, $F_1$-measure and the Accuracy in aspects of verifying the efficiency and accuracy of the proposed model. The mathematical formulas is listed as follow:
\begin{equation} \label{Eq:pre}
  {Precision=\frac{TP}{TP+FP}},
\end{equation}
\begin{equation} \label{Eq:recall}
  {Recall=\frac{TP}{TP+FN}},
\end{equation}
\begin{equation} \label{Eq:f1}
  {F_1=\frac{2Recall\times Precision}{Recall+Precision}},
\end{equation}
\begin{equation} \label{Eq:acc}
  {Accuracy=\frac{TP+FN}{TP+FN+TN+FP}},
\end{equation}
where $TP$ (true positive) denotes that the glomerulus with spike is correctly classified; $FP$ (false positive) represents that the glomerulus without spike is wrongly classified as the glomerulus with spike; $FN$ (false negative) indicates that the glomerulus with spike is wrongly classified as the glomerulus without spike; $TN$ (true negative) means that the glomerulus without spike is correctly classified.

Besides, we utilize the Receiver operating characteristic (ROC) curve to measure the sensitivity and specificity of the proposed model. For the coordinate of ROC curve, the horizontal and vertical axis represent false positive rate ($FPR$) and true positive rate ($TPR$), respectively. The specific formulations is given as follow:
\begin{equation} \label{Eq:tpr}
  {TPR=\frac{TP}{TP+FN}},
\end{equation}
\begin{equation} \label{Eq:fpr}
  {FPR=\frac{TP}{FP+TN}}.
\end{equation}

We can from the ROC curve calculate the AUC value, which is the area under the ROC curve and is a standard measuring the quality of the classification model. Generally, the value of AUC is between $0.5$ and $1.0$. The larger the AUC value, the better the model performance.

\subsection{Verification of classification efficiency}
\begin{figure}[htb]
\centerline{\includegraphics[width=5 in,height=4 in]{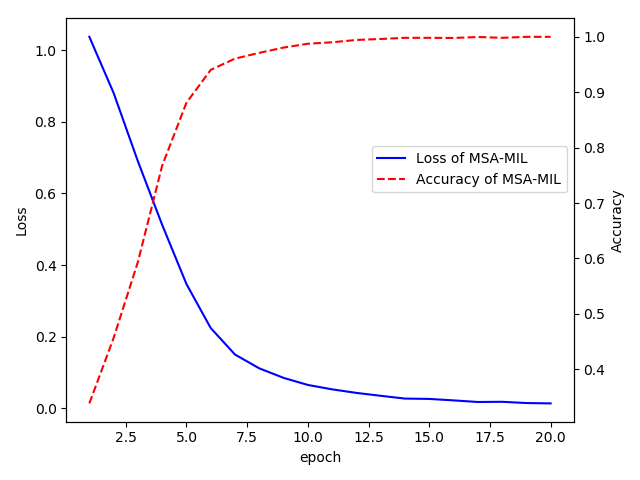}}
\caption{Curve of the loss function and accuracy in training the MSA-MIL model.}
\label{Loss and accuracy of MSA-MIL}
\end{figure}

After dividing the data set as training data and test data, we use the training set containing $1015$ glomerular images to train the MSA-MIL. Fig.\ref{Loss and accuracy of MSA-MIL} displays the change of the loss and accuracy with training epoch increasing. we can see that the training of the model is effective, and the model converges at the $10$th epoch with lower loss and higher accuracy.

\subsection{Visualization of glomerular spikes}
\label{visual}
We use a sliding window aggregation method to visualize the spikes of glomeruli. We choose the same window size as the tile size to classify the images in turn, and obtain the classification probability of each window. By connecting the probabilities of all tiles in an image, a probability distribution matrix can be formed. According to the probability distribution matrix, we can draw a heatmap which is superimposed on the original image with a certain transparency to obtain the visualization results of glomerular spikes. The heatmap can reflect the location of the spike in the glomerulus at a certain extent. It solves the unexplainable problem of the current classification network. At the same time, the doctor can understand the basis of the algorithm classification through the heatmap of the spike, so as to trust the result of the classification.

\begin{figure}[htb]
\centerline{\includegraphics[width=6.5 in,height=6.5 in]{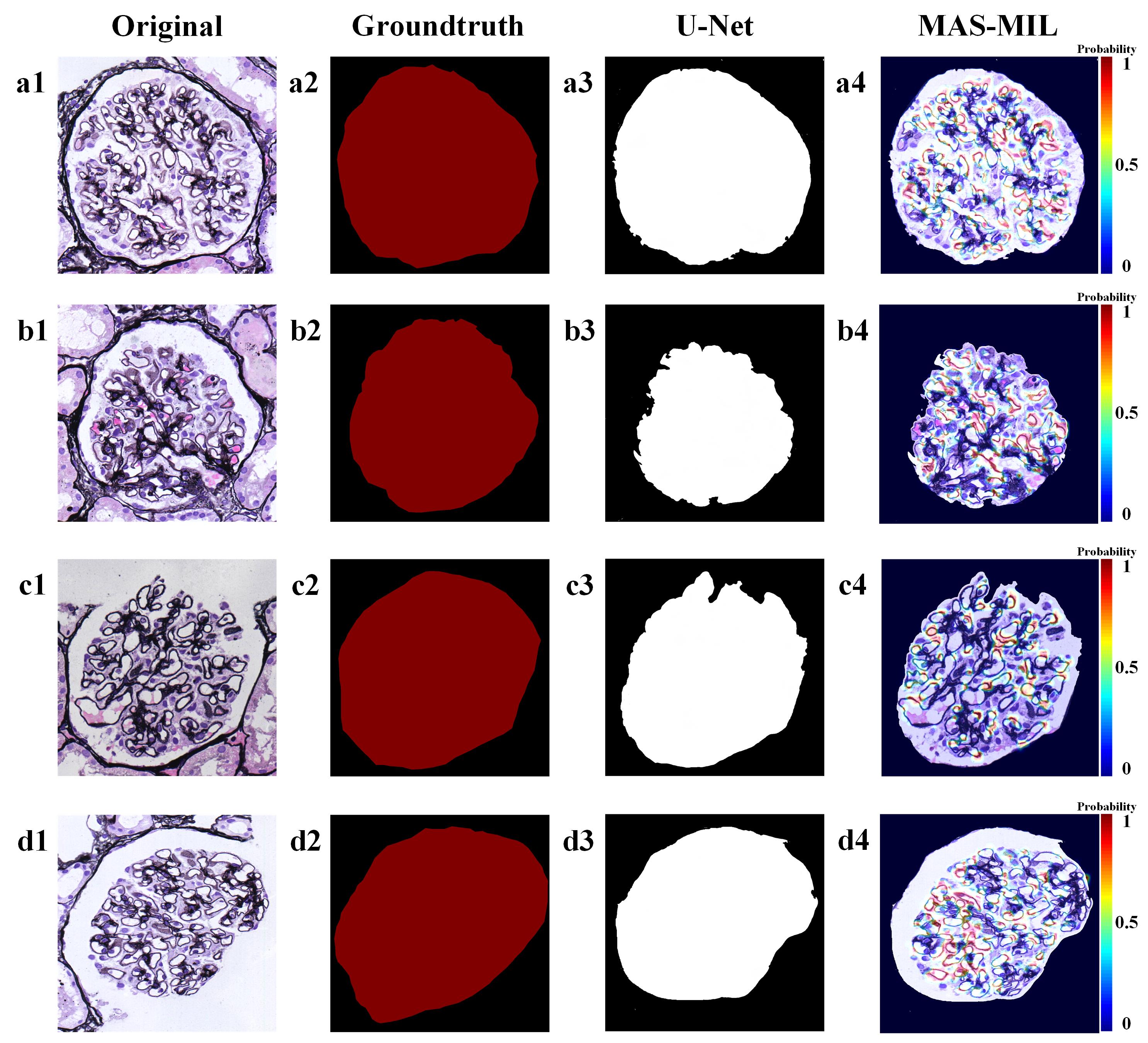}}
\caption{Spike visualization for four glomeruli images by the MSA-MIL model. First column: original image of the glomeruli; Second column: ground truth for training U-Net. Third column: segmentation results of U-Net. Fourth column: visualization of spike in the original image.}
\label{Visualization of spikes}
\end{figure}

Fig.\ref{Visualization of spikes} shows the visualized results of four glomerulus images by the proposed model. The first column displays four original images of the glomeruli, the second column shows the ground truth for training U-Net. the third column is corresponding region extracting results of glomerular through the U-Net network and the fourth column gives the visualization results of spikes in the corresponding original images. The color from blue to red in the regions of heatmap indicates that the probability being a spike. The blue area indicates that the area has a low probability of being spikes, and the red area indicates that the area has a high probability of being spikes. It can be seen from the last column of Fig.\ref{Visualization of spikes} that the MSA-MIL has achieved the goal of visualizing the location of the spikes in the glomeruli.

\subsection{Optimal parameters setting}
\label{parameters}
In order to obtain the best performance of the trained model, we need to choose the appropriate size of tile in dividing the image into many tiles and the ratio of positive and negative instances when updating the temporary training data set $D(x_d^{(ij,k)},y_d^{(ij,k)})$.

\begin{figure}[htb]
\centerline{\includegraphics[width=6 in,height=2.16 in]{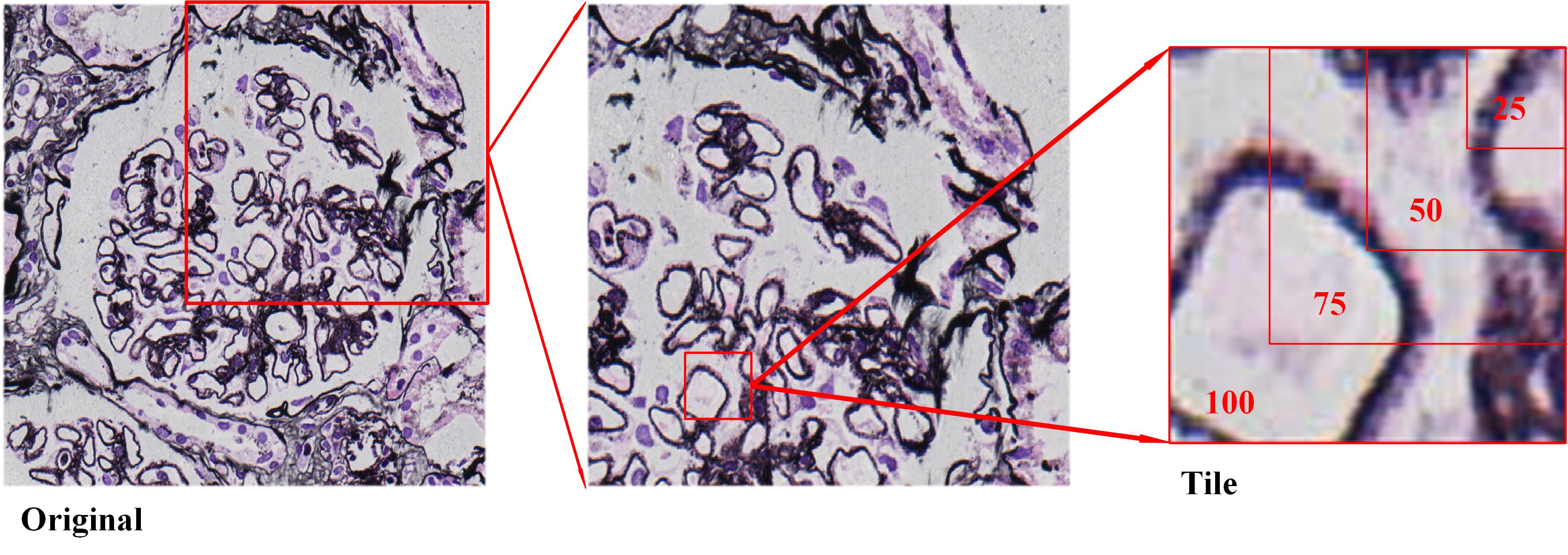}}
\caption{Different sizes (25, 50, 75, 100) of tile tested in the proposed model.}
\label{Schematic diagram of tile size}
\end{figure}

\begin{figure}[htb]
  \centering
  \subfigure[]{\includegraphics[width=3.in]{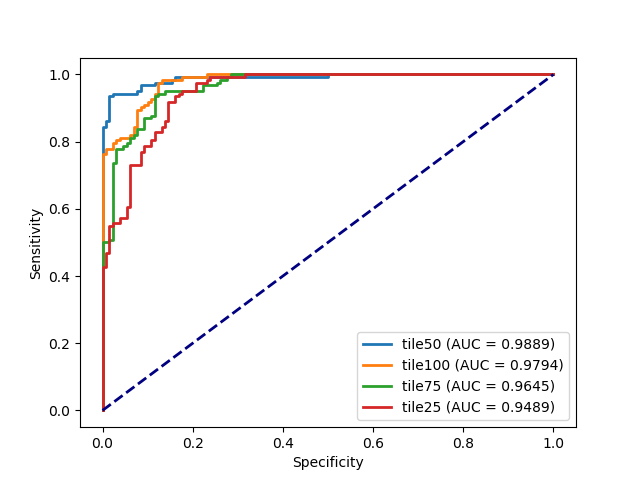}}
  \subfigure[]{\includegraphics[width=3.in]{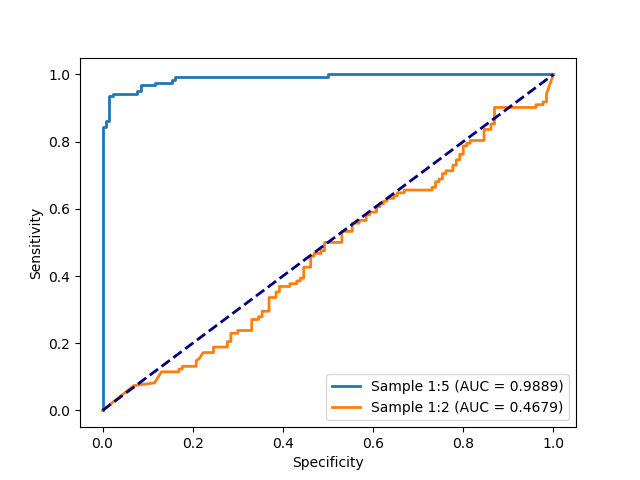}}
  \caption{ROC curve of different tile sizes and different sample proportion. (a) ROC curve of different tile sizes. (b) ROC curve of different sample proportion.}
  \label{ROC of para}
\end{figure}

The size of tile has a great influence on the accuracy of model learning. Larger tile size can not learn small features, while smaller tile size will cause the model to fail to learn the integrable features. In order to select the appropriate tile, we use different sizes of tile (25, 50, 75 and 100) to test the proposed model and calculate value of the four metrics (Precision, Recall, $F_1$-measure and  Accuracy) which are shown in Table\ref{tilesize}. The results show that the model achieve the best classification accuracy when the size of tile is set as $50\times50$. Fig.\ref{Schematic diagram of tile size} shows different sizes of tile used in the proposed model, from which we can observe that the tile with the size of $50$ just fits into the basement membrane with spikes, fine-grained features cannot be learned with the tile size exceeding $50$ and the features are too fine and lack of integrity with the tile size under $50$. The result can be confirmed by the AUC value shown in Fig.\ref{ROC of para}(a) in which the AUC value is the highest for setting the tile size as $50$.

\begin{table}[htbp]
  \centering
  \caption{Classification performance of different tile sizes}
  \label{tilesize}

  \begin{tabular}{lcccc}
    \toprule
    Tile size & Precision & Recall & F1-score & Accuracy \\
    \midrule
    25 & 0.8525 & 0.8525 & 0.8525 & 0.8571 \\
    \textbf{50} & \textbf{0.9828} & \textbf{0.9344} & \textbf{0.9580} & \textbf{0.9603} \\
    75 & 0.9592 & 0.7705 & 0.8545 & 0.8730 \\
    100 & 0.9429 & 0.8115 & 0.8722 & 0.8849 \\
    \bottomrule
  \end{tabular}
\end{table}

When training the instance-level network, the ratio of positive and negative instances updating the training set is a parameter that needs to be adjusted. According to the setting of MIL, MIL uses the label of the image as the label of tile and thus the tile in the positive bag may be mislabeled. When updating the temporary training data set $D(x_d^{(ij,k)},y_d^{(ij,k)})$, the positive instances with the highest probability in the positive bag are put into the training set at each iteration. At this point, the number of negative instances will affect the accuracy of the model. we test two ratios of positive instance and negative instance with $1:2$ and $1:5$ and calculate the same four metric. The results in Table \ref{proportion} show that the sufficient number of negative samples ensures that the model can learn the features of more negative instances, thus ensuring the accuracy of the classification results. The ROC curve shown in Figure.\ref{ROC of para}(b) also confirm that the AUC value arrive the highest by setting the ratio of positive and negative instances as $1:5$, that is, the positive instance with the highest probability is selected from the positive bag each time, and the negative instances with the top five probabilities from the negative bag are selected and put into the training set for training.

\begin{table}[htb]
  \centering
  \caption{Classification performance of different sample proportion}
  \label{proportion}

  \begin{tabular}{lcccc}
    \toprule
    Sample proportion & Precision & Recall & F1-score & Accuracy \\
    \midrule
    1:2 & 0.4933 & 0.9016 & 0.6377 & 0.5833 \\
    \textbf{1:5} & \textbf{0.9828} & \textbf{0.9344} & \textbf{0.9580} & \textbf{0.9603} \\
    \bottomrule
  \end{tabular}
\end{table}

\subsection{Comparison with competing models}
\label{comparison}

\begin{figure}[htbp]
\centerline{\includegraphics[width=5 in,height=4 in]{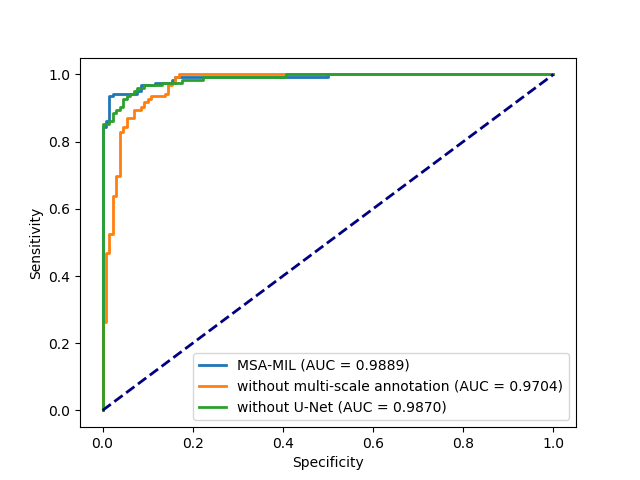}}
\caption{ROC curve of three competing models.}
\label{ROC of model}
\end{figure}

In order to verify the effectiveness of MSA-MIL in classification, we selected several models for comparison. They are the MIL model without multi-scale annotation, the model without U-Net structure and the ordinary ResNet-18 classification model. For the proposed model MSA-MIL, U-Net is a necessary part of the network structure to remove extraneous regions outside the glomeruli. The features learned by the algorithm are mainly concentrated on the glomeruli, which improves the efficiency of the model. In addition, a common ResNet-18 is trained to classify spiked glomeruli and non-spiked glomeruli. The value of four metric obtained from the competing model are listed in Table \ref{modelcomparison}. The results in Table \ref{modelcomparison} show that the multi-scale annotation training method proposed in this paper achieve higher accuracy. It indicates that the method of multi-scale labeling solves the problem of MIL label instability, and greatly improves the accuracy of classification. Furthermore, Fig.\ref{ROC of model} shows the ROC curves of the MSA-MIL, the model without multi-scale annotation and the model without U-Net. It can be seen from the ROC curve that the AUC value of the MSA-MIL is high and reaches $0.9889$. Therefore, the proposed model has better stability.

\begin{table}[htbp]
  \centering
  \caption{Comparison results of four competing models}
  \label{modelcomparison}

  \begin{tabular}{lcccc}
    \toprule
    Model & Precision & Recall & F1-score & Accuracy \\
    \midrule
    without U-Net & 0.9483 & 0.9016 & 0.9244 & 0.9286 \\
    without multi-scale annotation & 0.9217 & 0.8689 & 0.8945 & 0.9008 \\
    ResNet-18 & 0.8661 & 0.9016 & 0.8835 & 0.8849 \\
    \textbf{MSA-MIL} & \textbf{0.9828} & \textbf{0.9344} & \textbf{0.9580} & \textbf{0.9603} \\
    \bottomrule
  \end{tabular}
\end{table}

\subsection{Comparison with Pathologists}
\label{pathologists}
The proposed model is compared with pathologists in the classification performance of glomerulus and speed of marking the location of the spike. $252$ glomerular images in the test dataset were classified and labeled by a junior pathologist with three years of work experience and a senior pathologist with five years of work experience. They classify all the glomeruli and mark the location of the spike and record the time spent on processing each glomerulus. Table \ref{pathologistscomparison} shows the results of the average speed and performance of each glomerulus by the proposed model and two pathologists. As can be seen, the performance and speed of the pathologists are low and unstable because of subjective factors such as lack of clinical experience and work fatigue. Pathologists are prone to miss detection when classifying because the characteristics of the spikes are small and not obvious. MAS-MIL effectively helps medical pathologists to diagnose spikes and improves the efficiency of diagnosis.

\begin{table}[htbp]
  \centering
  \caption{Comparison results of MSA-MIL and two pathologists}
  \label{pathologistscomparison}

  \begin{tabular}{lccccc}
    \toprule
    Model & Time & Precision & Recall & F1-score & Accuracy \\
    \midrule
    Primary pathologist & 99.58s & \textbf{1.0000} & 0.3934 & 0.5647 & 0.7063 \\
    Senior pathologist & 87.32s & \textbf{1.0000} & 0.5902 & 0.7423 & 0.8016 \\
    MSA-MIL & \textbf{1.33s} & 0.9828 & \textbf{0.9344} & \textbf{0.9580} & \textbf{0.9603} \\
    \bottomrule
  \end{tabular}
\end{table}

\section{Conclusions}
\label{sec-conclus}
In this paper, we present a MSA-MIL method to classify and visualize spikes on the glomerular basement membrane. It combines U-Net and ResNet under the framework of multiple instance learning, and use multi-scale annotation data for training. We totally used $1267$ glomerular images as dataset. In the test group, the F-measures of the glomerular classification reached 0.9580. The experimental results show that this method can effectively classify the spiked glomerulus and normal glomerulus, and can give a visualization result of the position of the spike in the glomerulus. Compared with other methods and pathologists, this method can significantly improve the accuracy of classification and accelerate the efficiency of diagnosing glomerulus. Furthermore, the visualization of the lesion make the clinical doctors more convinced the classification results. Therefore, The proposed model provides a good foundation for clinical intelligent diagnosis of nephropathy and plays an important role in assisting the diagnosis of glomerular membranous nephropathy.

\section*{Conflict of interest}
\label{sec-conflict}
None declared.

\section*{Acknowledgment}
\label{sec-ack}
This work was supported by the National Natural Science Foundation of China (Grant NO. 11472184, No. 11771321, No. 61901292); the National Youth Science Foundation of China (Grant NO.11401423); the ShanXi province plan project on Science and Technology of social Development (Grant No. 201703D321032);  and the Natural Science Foundation of Shanxi Province, China (Grant No. 201901D211080)


\begin{thebibliography}{}
\expandafter\ifx\csname url\endcsname\relax
  \def\url#1{\texttt{#1}}\fi
\expandafter\ifx\csname urlprefix\endcsname\relax\def\urlprefix{URL }\fi
\expandafter\ifx\csname href\endcsname\relax
  \def\href#1#2{#2} \def\path#1{#1}\fi

\end{thebibliography}


\begin{thebibliography}{99}
{\footnotesize
\bibitem{Spanhol22016}
Spanhol, F.A., Oliveira, L.E., Petitjean, C., Heutte, L. (2016). Breast cancer histopathological image classification using Convolutional Neural Networks. 2016 International Joint Conference on Neural Networks (IJCNN), 2560-2567.
\bibitem{Litjens2016}
Litjens, G.J., Sánchez, C.I., Timofeeva, N., Hermsen, M., Nagtegaal, I.D., Kovacs, I., Kaa, C.H., Bult, P., Ginneken, B.V., Laak, J.V. (2016). Deep learning as a tool for increased accuracy and efficiency of histopathological diagnosis. Scientific Reports, 6.
\bibitem{Janowczyk2016}
Janowczyk, A., Madabhushi, A. (2016). Deep learning for digital pathology image analysis: A comprehensive tutorial with selected use cases. Journal of Pathology Informatics, 7.
\bibitem{Pedraza2017}
Pedraza, A., Gallego, J., Lopez, S., Gonzalez, L., Laurinavicius, A., Bueno, G. (2017). Glomerulus Classification with Convolutional Neural Networks. MIUA.
\bibitem{Simon2018}
Simon, O., Yacoub, R., Jain, S., Tomaszewski, J.E., Sarder, P. (2018). Multi-radial LBP Features as a Tool for Rapid Glomerular Detection and Assessment in Whole Slide Histopathology Images. Scientific Reports, 8.
\bibitem{Kawazoe2018}
Kawazoe, Y., Shimamoto, K., Yamaguchi, R., Shintani-Domoto, Y., Uozaki, H., Fukayama, M., Ohe, K. (2018). Faster R-CNN-Based Glomerular Detection in Multistained Human Whole Slide Images. J. Imaging, 4, 91.
\bibitem{LiuR2019}
Liu, R.H., Wang, L., He, J., Chen, W. (2019). Towards Staining Independent Segmentation of Glomerulus from Histopathological Images of Kidney.  bioRxiv, 2019: 821181.
\bibitem{Bueno2020}
Bueno, G., $Fernandez-Carrobles$, M., $Gonzalez-Lopez$, L., $Deniz-Suarez$, O. (2020). Glomerulosclerosis identification in whole slide images using semantic segmentation. Computer methods and programs in biomedicine, 184, 105273.
\bibitem{Barros2017}
Barros, G.O., Navarro, B., Duarte, A., dos-Santos, W.L. (2017). PathoSpotter-K: A computational tool for the automatic identification of glomerular lesions in histological images of kidneys. Scientific Reports, 7.
\bibitem{Marsh2018}
Marsh, J.N., Matlock, M.K., Kudose, S., Liu, T., Stappenbeck, T.S., Gaut, J.P., Swamidass, S.J. (2018). Deep Learning Global Glomerulosclerosis in Transplant Kidney Frozen Sections. IEEE Transactions on Medical Imaging, 37, 2718-2728.
\bibitem{Kannan2019}
Kannan, S., Morgan, L.A., Liang, B., Cheung, M.G., Lin, C.Q., Mun, D., Nader, R., Belghasem, M.E., Henderson, J.M., Francis, J., Chitalia, V.C., Kolachalama, V.B. (2019). Segmentation of Glomeruli Within Trichrome Images Using Deep Learning. Kidney International Reports, 4, 955 - 962.
\bibitem{Ginley2019}
Ginley, B., Lutnick, B., Jen, K., Fogo, A.B., Jain, S., Rosenberg, A.Z., Walavalkar, V., Wilding, G., Tomaszewski, J.E., Yacoub, R., Rossi, G.M., Sarder, P. (2019). Computational Segmentation and Classification of Diabetic Glomerulosclerosis. Journal of the American Society of Nephrology : JASN.
\bibitem{Uchino2020}
Uchino, E., Suzuki, K., Sato, N., Kojima, R., Tamada, Y., Hiragi, S., Yokoi, H., Yugami, N., Minamiguchi, S., Haga, H., Yanagita, M., Okuno, Y. (2020). Classification of glomerular pathological findings using deep learning and nephrologist-AI collective intelligence approach. medRxiv.
\bibitem{Chagas2020}
Chagas, P., Souza, L., Araújo, I., Aldeman, N., Duarte, A.A., Angelo, M., dos-Santos, W.L., Oliveira, L. (2020). Classification of glomerular hypercellularity using convolutional features and support vector machine. Artificial intelligence in medicine, 103, 101808
\bibitem{Shen2015}
Shen, W., Zhou, M., Yang, F., Yang, C., Tian, J. (2015). Multi-scale Convolutional Neural Networks for Lung Nodule Classification. Information processing in medical imaging : proceedings of the ... conference, 24, 588-99 .
\bibitem{Setio2016}
Setio, A.A., Ciompi, F., Litjens, G.J., Gerke, P.K., Jacobs, C., Riel, S.J., Wille, M.M., Naqibullah, M., Sánchez, C.I., Ginneken, B.V. (2016). Pulmonary Nodule Detection in CT Images: False Positive Reduction Using Multi-View Convolutional Networks. IEEE Transactions on Medical Imaging, 35, 1160-1169.
\bibitem{YuQ2019}
Yu, Q., Shi, Y., Sun, J., Gao, Y., Zhu, J., Dai, Y. (2019). Crossbar-Net: A Novel Convolutional Neural Network for Kidney Tumor Segmentation in CT Images. IEEE Transactions on Image Processing, 28, 4060-4074.
\bibitem{Dou2016}
Dou, Q., Chen, H., Yu, L., Zhao, L., Qin, J., Wang, D., Mok, V.C., Shi, L., Heng, P. (2016). Automatic Detection of Cerebral Microbleeds From MR Images via 3D Convolutional Neural Networks. IEEE Transactions on Medical Imaging, 35, 1182-1195.
\bibitem{Yang2015}
Yang, D., Zhang, S., Yan, Z., Tan, C., Li, K., Metaxas, D.N. (2015). Automated anatomical landmark detection ondistal femur surface using convolutional neural network. 2015 IEEE 12th International Symposium on Biomedical Imaging (ISBI), 17-21.
\bibitem{Aldoj2019}
Aldoj, N., Lukas, S., Dewey, M., Penzkofer, T. (2019). Semi-automatic classification of prostate cancer on multi-parametric MR imaging using a multi-channel 3D convolutional neural network. European Radiology, 30, 1243-1253.
\bibitem{Gulshan2016}
Gulshan, V., Peng, L.H., Coram, M., Stumpe, M.C., Wu, D.J., Narayanaswamy, A., Venugopalan, S., Widner, K., Madams, T., Cuadros, J., Kim, R., Raman, R., Nelson, P.C., Mega, J.L., Webster, D.R. (2016). Development and Validation of a Deep Learning Algorithm for Detection of Diabetic Retinopathy in Retinal Fundus Photographs. JAMA, 316 22, 2402-2410.
\bibitem{Esteva2017}
Esteva, A., Kuprel, B., Novoa, R.A., Ko, J., Swetter, S.M., Blau, H.M., Thrun, S. (2017). Dermatologist-level classification of skin cancer with deep neural networks. Nature, 542, 115-118.
\bibitem{Spanhol2016}
Spanhol, F.A., Oliveira, L.E., Petitjean, C., Heutte, L. (2016). Breast cancer histopathological image classification using Convolutional Neural Networks. 2016 International Joint Conference on Neural Networks (IJCNN), 2560-2567.
\bibitem{Wei2017}
Wei, B., Han, Z., He, X., Yin, Y. (2017). Deep learning model based breast cancer histopathological image classification. 2017 IEEE 2nd International Conference on Cloud Computing and Big Data Analysis (ICCCBDA), 348-353.
\bibitem{He2016}
He, K., Zhang, X., Ren, S., Sun, J. (2016). Deep Residual Learning for Image Recognition. 2016 IEEE Conference on Computer Vision and Pattern Recognition (CVPR), 770-778.
\bibitem{Ronneberger2015}
Ronneberger, O., Fischer, P., Brox, T. (2015). U-Net: Convolutional Networks for Biomedical Image Segmentation. ArXiv, abs/1505.04597.
\bibitem{Cao2019}
Cao, H., Liu, H., Song, E., Ma, G., Xu, X., Jin, R., Liu, T., Hung, C. (2019). Two-Stage Convolutional Neural Network Architecture for Lung Nodule Detection. ArXiv, abs/1905.03445.
\bibitem{Xian2018}
Xian-cheng, W., Wei, L., Bingyi, M., He, J., Jiang, Z., Xu, W.C., Ji, Z., Hong, G., Zhaomeng, S. (2018). International Conference on Data Science (ICDS 2018) Retina Blood Vessel Segmentation Using A U-Net Based Convolutional Neural Network.
\bibitem{Dietterich1997}
Dietterich, T.G., Lathrop, R.H., Lozano-Pérez, T. (1997). Solving the Multiple Instance Problem with Axis-Parallel Rectangles. Artif. Intell., 89, 31-71.
\bibitem{Maron1997}
Maron, Oded Lozano-Pérez, Tomás. (1997). A Framework for Multiple-Instance Learning. Advances in neural information processing systems. 10.
\bibitem{Maron1998}
Maron, O., Ratan, A.L. (1998). Multiple-Instance Learning for Natural Scene Classification. ICML.
\bibitem{Yang2000}
Yang, C., Lozano-Perez, T. (2000). Image database retrieval with multiple-instance learning techniques. Proceedings of 16th International Conference on Data Engineering (Cat. No.00CB37073), 233-243.Parallel Rectangles. Artif. Intell., 89, 31-71.
\bibitem{Xu2016}
Xu, Y. (2016). Multiple-instance learning based decision neural networks for image retrieval and classification. Neurocomputing, 171, 826-836.
\bibitem{Xu2014}
Xu, Y., Mo, T., Feng, Q., Zhong, P., Lai, M., Chang, E.I. (2014). Deep learning of feature representation with multiple instance learning for medical image analysis. 2014 IEEE International Conference on Acoustics, Speech and Signal Processing (ICASSP), 1626-1630.
\bibitem{Hou2015}
Hou, L., Samaras, D., Kurç, T.M., Gao, Y., Davis, J.E., Saltz, J.H. (2015). Efficient Multiple Instance Convolutional Neural Networks for Gigapixel Resolution Image Classification. ArXiv, abs/1504.07947.
\bibitem{Campanella2019}
Campanella, G., Hanna, M.G., Geneslaw, L., Miraflor, A.P., Silva, V.W., Busam, K.J., Brogi, E., Reuter, V.E., Klimstra, D.S., Fuchs, T.J. (2019). Clinical-grade computational pathology using weakly supervised deep learning on whole slide images. Nature Medicine, 1-9.
}
\end{thebibliography}
\end{document}